# A High-Performance HOG Extractor on FPGA

Vinh Ngo
Department of Microelectronics and Electronics Systems
Spain
quangvinh.ngo@uab.cat

Arnau Casadevall
Department of Microelectronics and Electronics Systems
Spain
arnau.casadevall@uab.cat

Marc Codina
Department of Microelectronics and Electronics Systems
Spain
marc.codina@uab.cat

David Castells-Rufas
Department of Microelectronics and Electronics Systems
Spain
david.castells@uab.cat

Jordi Carrabina
Department of Microelectronics and Electronics Systems
Spain
jordi.carrabina@uab.cat

## ABSTRACT

Pedestrian detection is one of the key problems in emerging self-driving car industry. And HOG algorithm has proven to provide good accuracy for pedestrian detection. There are plenty of research works have been done in accelerating HOG algorithm on FPGA because of its low-power and high-throughput characteristics. In this paper, we present a high-performance HOG architecture for pedestrian detection on a low-cost FPGA platform. It achieves a maximum throughput of 526 FPS with 640x480 input images, which is 3.25 times faster than the state of the art design. The accelerator is integrated with SVM-based prediction in realizing a pedestrian detection system. And the power consumption of the whole system is comparable with the best existing implementations.

## 1 INTRODUCTION

Pedestrian detection is a safety critical application on autonomous cars. There are two main approaches to implement pedestrian detection systems. On one hand, the detection algorithm relies on all input image pixels. This approach uses deep learning method and it requires costly computing platforms with not only many processing cores but also large memory bandwidth and capacity. On the other hand, only extracted features from the image are input to the detection algorithm. This approach using HOG (Histogram of Gradients) [1] has proven to have good accuracy in detection [2]. While requiring less memory capacity, it is still a computing-intensive algorithm, which needs a low latency and high-throughput platform. FPGA, therefore, comes as suitable solution thanks to its capability in parallel processing. More importantly, FPGAs potentially have better energy efficiency in comparison with alternative platforms such as CPU and GPU.

In this paper, we design and implement a hog feature extractor on a low-cost FPGA device, targeting at high throughput and low power consumption. This work is based on our previous work in [3]. There are several improvements to help achieve a high-performance design. First, the fixed-point number is used to represent values other than the integer number, which apparently increases the feature's accuracy with the cost of computational complexity. Secondly, a pipeline for normalizing cell features to take advantages of hardware's capability in pipeline and parallel execution. The output HOG normalized features are transferred to the HPS (Hard Processor System) for prediction process. Third, instead of buffering input images before extracting, which costs memory, input pixels are processed directly from the sensor by a pipeline. And finally, we optimize the pipeline design so as to achieve the highest throughput. The HOG extractor design can work at a maximum clock frequency of 162 MHz and provide a throughput of 526 FPS, the highest throughput in the state of the art. The design is then integrated into a heterogeneous system with SVM-based prediction software. The energy efficiency is comparable to the most efficient implementations.

The paper is outlined as follows. An overview of the original HOG algorithm is described in section 2. Section 3 discusses related works regarding FPGA implementations of real-time HOG extractor. Section 4 presents our architectural design in detail. The experimental results and discussions are shown in section 5. Finally, the conclusions are presented in section 6.

## 2 HOG OVERVIEW

The HOG algorithm consists of two main steps: gradient computation and histogram generation.

To compute the gradient of a *pixel (x,y)*, first, we need to calculate the intensity difference of its two pairs of neighbor pixels in horizontal and vertical directions following the Equation (1) and (2) respectively.

$$Gx(x,y) = I(x+1,y) - I(x-1,y) \quad (1)$$
$$Gy(x,y) = I(x,y+1) - I(x,y-1) \quad (2)$$

Then, the magnitude and the orientation of the gradient at *pixel(x,y)* are computed by Equation (3) and (4).

$$|G(x,y)| = \sqrt{Gx(x,y)^2 + Gy(x,y)^2} \quad (3)$$
$$\phi(x,y) = arctan\left(\frac{Gy(x,y)}{Gx(x,y)}\right) \quad (4)$$

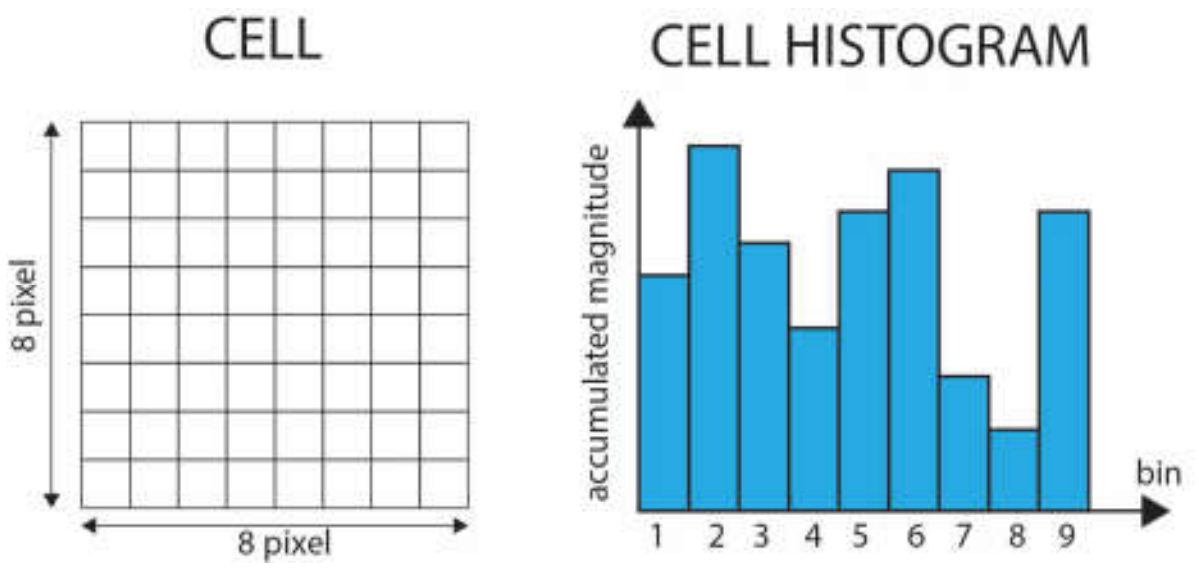

**Figure 1: Histogram is generated cell by cell**

Having the gradients, the histogram is generated cell by cell. Each cell has a size of 8x8 pixels. Therefore, a cell consists of 64 pairs of magnitude and orientation gradient values. Depending on the associated orientations, the magnitude gradients are accumulated to the corresponding bins. A cell histogram with nine bins is illustrated in Figure 1. Figure 2 describes in detail how the orientation of the gradient is quantized into a range of 9 bins using the scale from 0 to 180°.

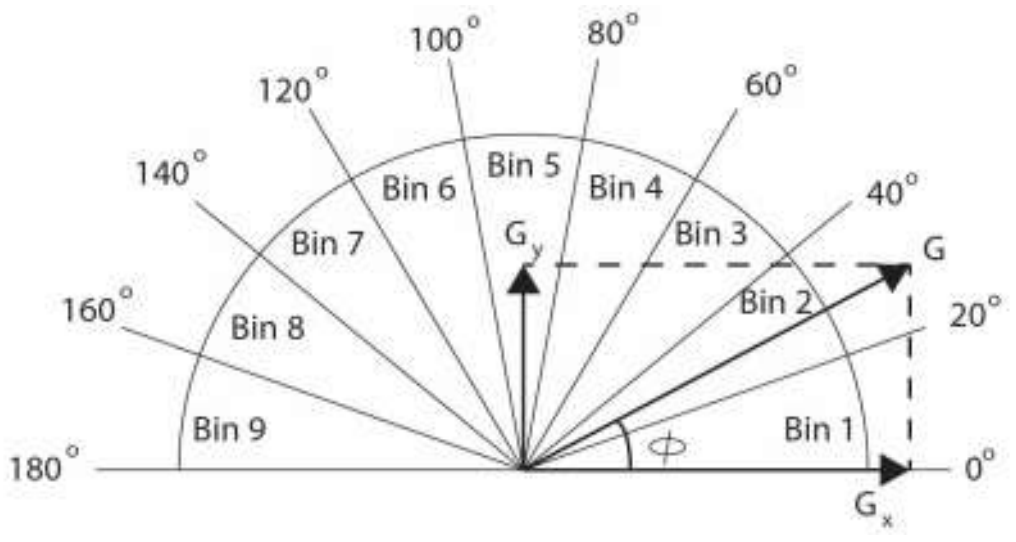

**Figure 2. Dividing into 9 bins from 0 to 180°**

The magnitude G, in this example, should be accumulated to bin 2 because its orientation is approximately 30°. For more accuracy, G will be accumulated fairly between adjacent bins depending on its exact orientation.

## 3 RELATED WORKS

To our knowledge, the works in [4], [5] presented the first implementations of HOG extractors on FPGAs. In [4], the HOG extractor is shown to have a good latency of only 312 µs. However, this design does not include the normalization module and it simplifies the computational process by using integer numbers. In [6], the authors proposed to process the pixel data at twice the pixel frequency and normalize the block histograms using L1-norm so that the available resources are efficiently used and can address parallel computing of multiple scales. With an input image of 1920x1080, the design achieves high speed with a latency of only 150 µs. But it is not clarified in the paper what this latency is about. Similarly, the design used some kinds of frame buffer before HOG processing module, which costs memory. Energy consumption of a HOG-based detection system on FPGA is first reported in [2]. In this work, the authors try to reduce the bit-width of the fixed-point representation to boost the performance. With a 640x480 frame size and a 13-bit fixed-point representation, the energy efficiency of the HOG extractor module is 0.54J/Frame. Anyway, the design leverages a costly hardware system with four FPGA devices and each device has 16 64-bit memory channels. The memory space for those 4 FPGA devices is 128 GB.

Another approach is presented in [7], in which the authors investigate the cell size and number of histogram bins that provide better performance. In this implementation, all the process of the detection system is integrated into an FPGA device. With a negligible loss in accuracy, the best set of parameters provides a frame rate of 42.7fps and high energy-efficiency of only 0.451J/Frame. A detailed description of HOG implementation on FPGA is presented in [8], which achieves a high processing speed at 40fps, with 1920x1080 input image size. Interestingly, in [9], HOG algorithm is analyzed on a heterogeneous system, including CPU, GPU, and FPGA. Based on multiple configuration experiments, the authors concluded that FPGA is best suited for histogram extraction and classification tasks in the whole detection flow because it produces a good trade-off between power and speed.

Recently, our previous work is published in [3], which simplifies the computing by using integer numbers. We achieved high throughput in HOG extracting process by buffering the input image. Besides, a look-up table is used to store the results of the square root and arctan computations. This approach heavily consumes on-chip memory. A low-complexity implementation of HOG-based pedestrian detection is presented recently in [10]. Instead of the original HOG, the authors proposed the use of histogram of significant gradients, and the hardware is, therefore, less complex. In addition, hardware resource usage is optimized by reducing the number of bits representing the intermediate values during computation processes. Besides, the authors avoid using complex representation numbers as well as DSP operations by pre-calculated values and simplification techniques.

## 4 IMPLEMENTATION

We implement the whole system in Terasic's DE1-SOC board. The system block diagram is shown in Figure 3. It includes the hardware components such as the image sensor, the HOG pipeline, the Hard Processor System (HPS), and other supporting modules.

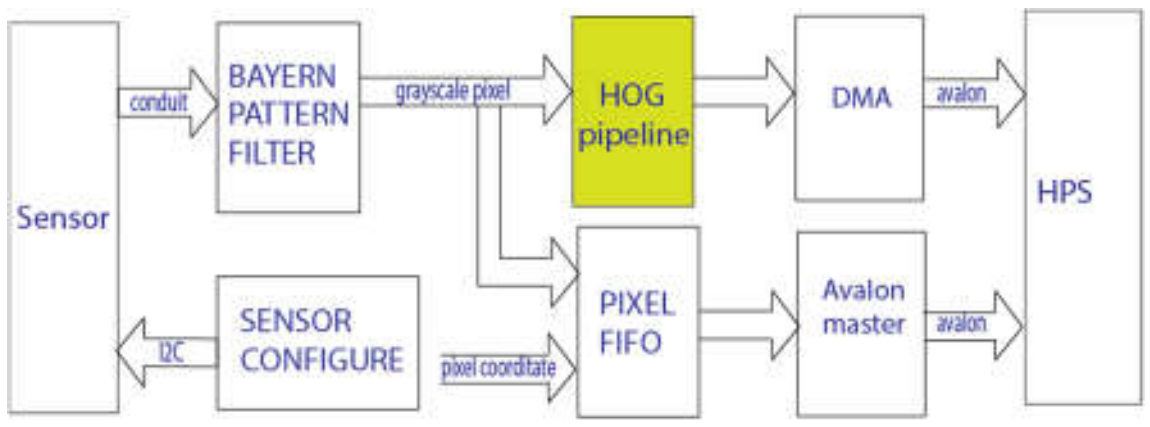

**Figure 3. System diagram**

Images from the sensor, after being filtered by the Bayern Pattern, are transferred directly to both the HOG pipeline module and the pixel FIFO. The pixel FIFO is necessary for later showing the original image on the VGA. A custom Avalon master interface is created to get pixels from this FIFO and write to the 1GB external SDRAM controlled by the HPS.

The image sensor is configured through an I2C interface for some key parameters such as image size, pixel clock. The Bayer pattern filter module takes raw input pixels and calculates the three colors pixel values. After that, the grayscale pixel value is generated to provide the HOG extractor and the HPS for real-time visualization.

The HOG extractor module is a long pipeline that generates the normalized hog features. Our best implementation in throughput used a 155 stages pipeline. The features are then written to the HPS memory by a DMA (Direct Memory Access). A Python code running on HPS will read these features out for predicting the present of pedestrians. The detailed architecture inside the HOG pipeline is presented in Figure 4.

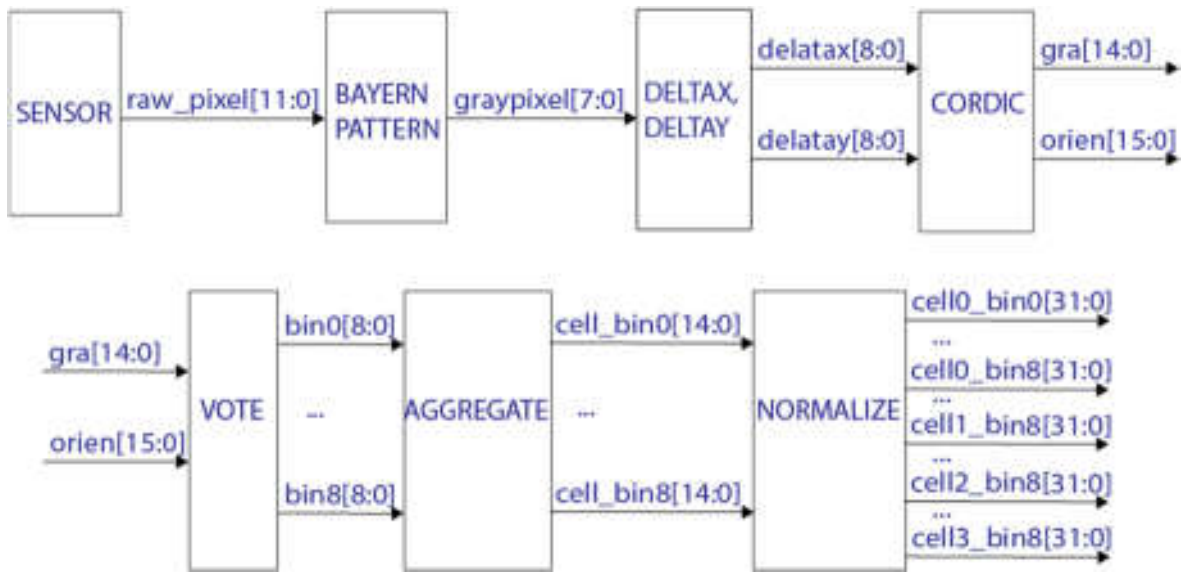

**Figure 4. HOG extractor block diagram**

First of all, luminance differences $G_x$ and $G_y$ (Eq. 1,2) are calculated by the DELTAXY module. These are 9 bit signed integers. We used the vector translate function in CORDIC IP to compute the magnitude and the orientation gradients. Both of them are fixed-point numbers. To achieve 2 digits after the decimal point accuracy, we choose to represent the orientation gradient by 13 fractional bits. Thus, the number of fractional bits for the magnitude gradient is six, according to the configuring requirement of CORDIC IP. Depending on the orientation gradient, the magnitude gradient of each pixel will vote to appropriate bins. The AGGREGATE module adds 64 histogram values of 64 pixels in a cell bin by bin to output the final cell features. Finally, cell features are block-wise contrast normalized. In this design, each block has four cells and L2 normalization [1] is chosen for the sake of accuracy and simplicity.

Figure 5 describes our hardware line buffers that allow the HOG module to compute the luminance difference $G_x$ and $G_y$ between neighbor pixels in vertical and horizontal directions. This design supports processing pixels on every clock cycle, which means that the performance of the design can be boosted if input pixels come at every clock cycle.

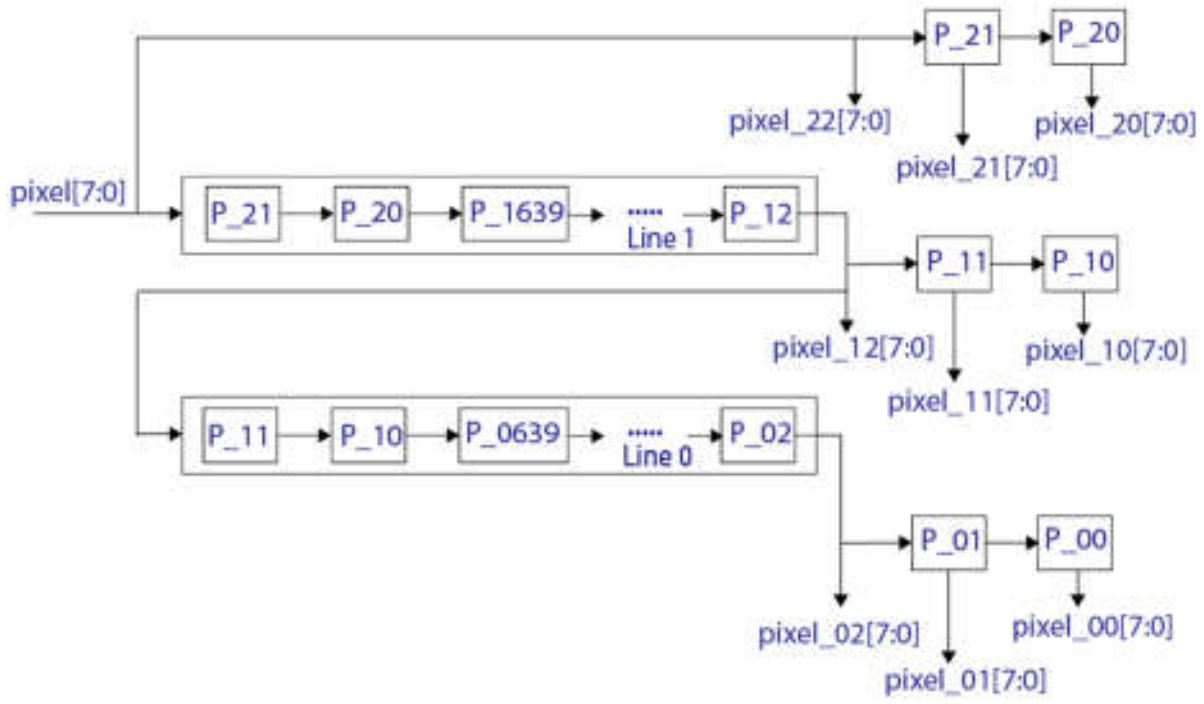

**Figure 5. Pixel line buffers**

The depth of each buffer corresponds to the row size of the input image, in our case 640. The luminance differences, $G_x$ and $G_y$, at pixel *P_11* are calculated using *P_21* and *P_01* for the vertical direction, and *P_10* and *P_12* for horizontal direction as in Equation (5) and (6).

$$Gx(1,1) = P_{10} - P_{12} \quad (5)$$
$$Gy(1,1) = P_{01} - P_{21} \quad (6)$$

Following the original HOG algorithm in [1], the final HOG feature is extracted from every cell of 8x8 pixel size. And the orientation is divided into 9 bins from 0 to 180º. In our case, with the 640x480 image size, the final HOG feature is a vector of 80x60x9 dimension. The HOG module processes in a pipeline approach every 8 continuous pixels in a row of a cell. And it generates a partial hog vector with 9 bins aggregating 8 magnitude gradient values. These partial hog vectors are put in a line buffer as shown in Figure 6. Only 80 partial cell hogs are needed to be stored so as to minimize the memory usage without stalling the pipeline. In order to generate the full HOG feature for a cell, it is necessary to aggregate 8 partial cell hogs from 8 different rows. The *cell_hog_valid* signal will be active only if all the partial cell hogs are fully collected.

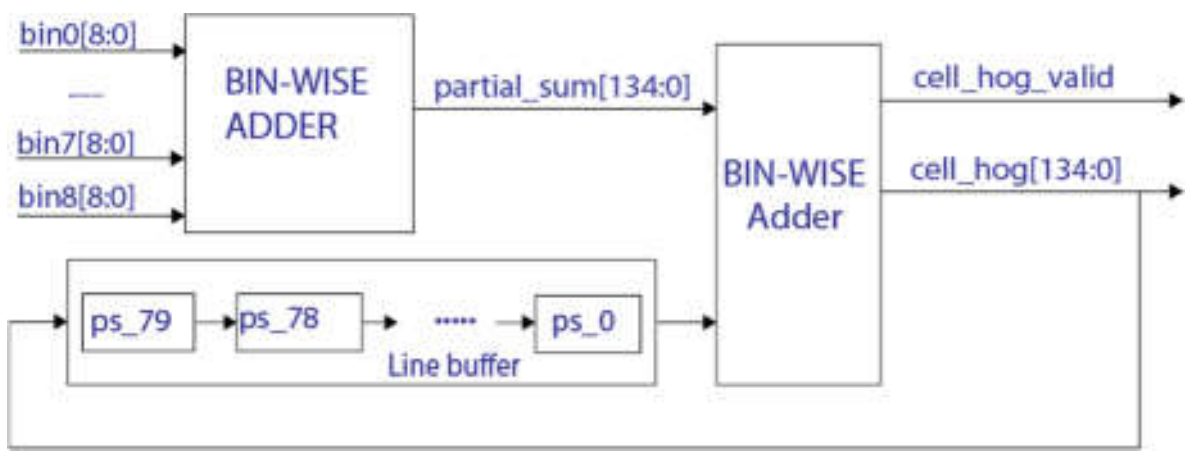

**Figure 6. Partial cell hog line buffer**

The normalization of the cell histogram is done following the equation in (7). In the equation, v is the cell hog features in the block, and ‖v‖2 is the L2-normalization of all the cell hog features in the block. A small constant, Ɛ, is added to avoid dividing by zero.

As illustrated in Figure 4, each bin of the normalized hog feature is represented by 32 bits. This is not the final HOG feature value and it is represented in floating-point format. We only do the conversion from the fixed-point format to the floating-point format for the final step. All the intermediate results are calculated by integer and fixed-point numbers depending on the specific task.

$$v = \frac{v}{\sqrt{\|v\|_2^2 + \varepsilon^2}} \qquad (7)$$

We used the ModelSim simulator and a C golden model of the HOG to verify the HOG extractor design. At the system level, we build a heterogeneous system as in Figure 7. An Ubuntu distribution runs on the HPS. We used Qsys (Quartus-II) to create the system-on-chip FPGA architecture.

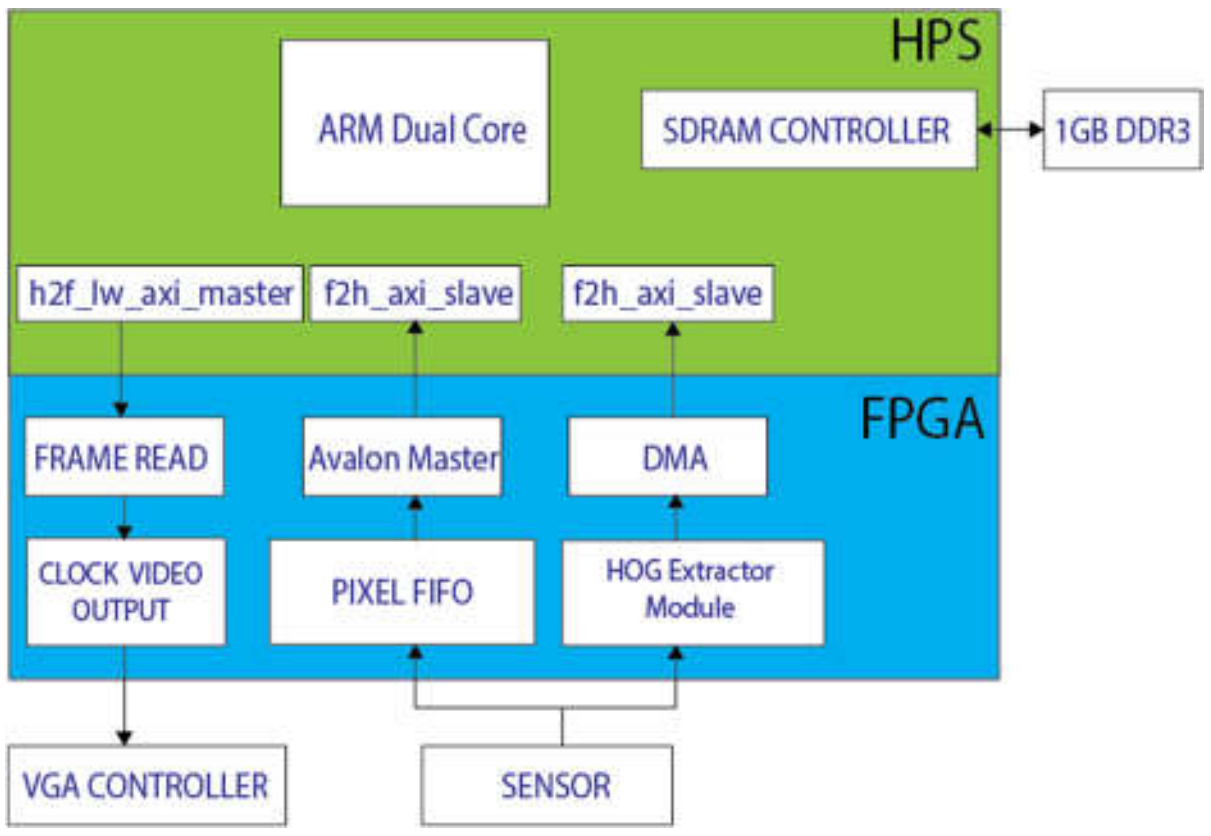

**Figure 7. HOG feature extraction system**

The normalized HOG feature is written to the external DDR3 SDRAM memory by a DMA through the *f2h_axi_slave* bridge. A custom Avalon bus master is created in Qsys to send image pixels also to the DDR3 memory. These two memory locations are set to be dedicated for FPGA. These transfer methods provide good system performance because data are transmitted in parallel with the HPS's CPU execution.

In the opposite direction, the pixels in the memory and detection results are sent to the VGA controller to visualize in real time.

To extract all the data (pixel image and hog features) we used a Python C/C++ API to write an extension module for binding the communicating between the HPS memory and the Python interpreter. That is useful to compute, in a high-level manner, the detection and classification tasks using Machine Learning techniques.

Our Python script is capable of reading both the frame and HOG vector provided by the board by reading in C their respective memory registers and, then, binding them with Python C/C++ API as Numpy [11] arrays.

We have trained and tested several models with different configurations using the INRIA Person Dataset [12] to achieve a maximum yield in terms of accuracy. We have tested our model not only with INRIA test dataset but also images from the camera sensor to have a more generalized model.

## 5 RESULTS AND DISCUSSIONS

The hardware design of the HOG extractor is compared step-wise with a reference C model which uses floating point operations. And the output results of the two implementations, which is the final normalized hog features, have an average difference of only 2 units at the second digit after the decimal point, which corresponds to 2% accurate because the range of a normalized feature value is from 0 to 1.

Table 1 reports the key compilation results for the HOG extractor and the heterogeneous system. For the HOG extractor, we provide two versions. The normal version works at 49 MHz and costs fewer hardware resources. The optimized one targets high throughput applications. It can work at 162 MHz clock frequency. Running at this frequency, the design only requires 2% more in ALMs and 9 more DSP blocks. And the number of registers is increased by nearly 20%. The on-chip memory usage is almost the same between the two versions. On the other hand, the heterogeneous system, as illustrated in Figure 7, is a system on chip design which includes the HOG extractor module as a hardware accelerator.

**Table 1: Compilation report for Cyclone V device**

| Design | Block memory Kbits | Logic (in ALMs) | DSP blocks | Registers | Fmax (MHz) |
|---|---|---|---|---|---|
| HOG extractor | 324 (8%) | 7,922 (25%) | 65 (75%) | 14,787 | 49 |
| Optimized HOG extractor | 326 (8%) | 8,610 (27%) | 74 (85%) | 17,697 | 162 |
| Heterogeneous system | 437 (11%) | 12,138 (38%) | 65 (75%) | 21,715 | 69 |

Other than the clock frequency, a design's output throughput also depends on the input throughput. Our design supports up to one input pixel every clock cycle. It means that if the input pixel clock is 162 MHz, our design throughput reaches 526 FPS as shown in Table 2. It is worth noting that the figures in this table are based on the HOG extractor design, not the heterogeneous system.

According to Table 2, our design speeds up 3.25x over the equivalent design recently published [10]. In order to compare designs that have different input frame size, we use pixels per second unit. In this scale, our design also supports the highest performance. Our implementation achieves 22% higher than the work in [6]. Another interesting measurement is the number of pixels per clock period. This number reflects the throughput of the design without taking into account the clock frequency. Our design achieves nearly 1 pixel on every clock, and it is the highest throughput according to Table 2.

Regarding FPGA resource, our design is optimized for memory usage and therefore consumes least memory resource among the designs in Table 2. The reason for this is that our pipeline works on every input pixel and there is not any buffer for input frames. For other resources, implementation in [6] is the most efficient. It targeted at low resource utilization by simplifying some computational operations. In the voting part, magnitudes are voted to only one unique bin without interpolation. Furthermore, all the calculations use integer numbers. On the other hand, the implementation in [13] is quite equivalent to ours. It consumes less LUTs, DSPs and registers thanks to a Look-up Table for storing in arctan values.

In terms of energy efficiency, we measure for the entire detection system. Following Table 3, our design consumes 0.82J/Frame which is quite higher than results from [2] and [7]. Those are the best designs in the state-of-the-art in terms of energy efficiency. The interesting point is that our design is better in power consumption. Thus, the reason for lower energy efficiency is because of the system detection rate. Our system only supports 11 fps although the HOG extractor can work at multiple times higher speed. We believe that our system throughput will improve significantly if we implement the classification task on FPGA, and therefore the energy consumption is reduced accordingly. We observed that sliding detection window, which is the most time-consuming task in the classification process, can be done in parallel in hardware but not in HPS's software. And we still have a large room to implement a classifier in hardware since the latency of the hardware part is only 6.16 ms, which is 14% of the design in [2]. Last but not least, Table 3 shows that our design's hardware resource is significantly small in comparison to the others.

**Table 2: Comparison of performance between different FPGA implementations**

| Design | Frame size | FPGA | Max frequency (MHz) | FPS | Pixels per Second (FPS*Frame Size) | Pixels per clock period (Pixels per Second/Frequency) | FPGA resources | | | |
|---|---|---|---|---|---|---|---|---|---|---|
| | | | | | | | Memory (Kb) | LUTs | DSPs | Registers |
| [4] | 800x600 | Spartan 3 | 63 | 30 | 14,400,000 | 0.229 | 1080 | 42,435 | - | - |
| [6] | 1920x1080 | Virtex 7 | 266 | 64 | 132,710,400 | 0.5 | 936 | 3,924 | 12 | 3,642 |
| [10] | 640x480 | Cyclone IV | 117.8 | 162 | 49,766,400 | 0.422 | - | - | - | - |
| [13] | 1920x1080 | ZynQ 7000 | 125 | 60 | 124,416,000 | 0.995 | 432 | 7,226 | 26 | 12,462 |
| Ours | 640x480 | Cyclone V | 162 | 526 | 161,587,200 | 0.997 | 326 | 8,610 | 74 | 17,697 |

**Table 3: Comparison of energy efficiency**

| | Frame size | FPGA | Freq. (MHz) | Latency | Power (W) | Energy (J/Frame) | FPS | Memory (Kb) | # of LUTs | DSPs | # of FFs |
|---|---|---|---|---|---|---|---|---|---|---|---|
| [2] | 640x480 | Virtex 6 | 150 | 44 ms | 37 | 0.54 | 68.2 | 13,738 | 184,953 | 190 | 208,666 |
| [7] | 1920x1080 | Virtex 7 | 266 | - | 19 | 0.451 | 42.7 | 4,079 | 30,360 | 364 | 48,576 |
| Us | 640x480 | Cyclone V | 50 | 6.16 ms | 9 | 0.82 | 11 | 437 | 12,138 | 65 | 21,715 |

## 6 CONCLUSION

A high-performance HOG feature extractor is implemented on a low-cost FPGA device. Fixed-point representation is employed for achieving approximately 2% different in comparison with the floating-point golden model. The HOG extractor design, which supports 526 FPS, can be a well-fitted IP in high-performance pedestrian detection systems.

The energy consumption of the whole detection system is 0.82J/Frame, which is among the good existing solutions. However, there is still room for future improvements to increase the detection throughput of our system design and lower down the energy consumption accordingly. This is can be done by implementing a classifier on chip. Thus, the classification task can slide the detection window through the HOG frame in parallel. Furthermore, the classification task also can be started early and pipelined together with the hog feature calculating process.

## ACKNOWLEDGMENTS

This work was supported by Spanish projects TEC2014-59679-C2-2.